\documentclass[review]{elsarticle}

\usepackage{lineno,hyperref}
\usepackage{url}
\usepackage{amssymb}
\usepackage{subfig}
\usepackage{amsmath}
\usepackage{mathtools}
\usepackage{color}
\usepackage{xcolor}
\usepackage{multirow}
\modulolinenumbers[5]

\journal{Journal of Visual Communication and Image Representation}

\begin{document}

\begin{frontmatter}

\title{Localized Trajectories for 2D and 3D Action Recognition}

\author[ad]{Konstantinos Papadopoulos \corref{mycorrespondingauthor}}
\cortext[mycorrespondingauthor]{Corresponding author}
\ead{konstantinos.papadopoulos@uni.lu}
\author[ad]{Girum Demisse}
\author[ad]{Enjie Ghorbel}

\author[ad1]{Michel Antunes}

\author[ad]{Djamila Aouada}
\author[ad]{Bj\"{o}rn Ottersten}

\address[ad]{Interdisciplinary Center for Security, Reliability and Trust, University of Luxembourg, Luxembourg}
\address[ad1]{Perceive3D, Coimbra Area, Portugal}





\begin{abstract}
The \textit{Dense Trajectories} concept is one of the most successful approaches in action recognition, suitable for scenarios involving a significant amount of motion. However, due to noise and background motion, many generated trajectories are irrelevant to the actual human activity and can potentially lead to performance degradation. In this paper, we propose \textit{Localized Trajectories} as an improved version of Dense Trajectories where motion trajectories are clustered around human body joints provided by RGB-D cameras and then encoded by local Bag-of-Words. As a result, the Localized Trajectories concept provides a more discriminative representation of actions as compared to Dense Trajectories. Moreover, we generalize Localized Trajectories to 3D by using the modalities offered by RGB-D cameras. One of the main advantages of using RGB-D data to generate trajectories is that they include radial displacements that are perpendicular to the image plane. Extensive experiments and analysis are carried out on five different datasets.
\end{abstract}


\end{frontmatter}

\section{Introduction}
\label{sec:introduction}

Human action recognition is an active research topic with several applications in surveillance and security~\cite{baptista2018anticipating}, healthcare and assisted living~\cite{baptista2017flexible, baptista2017video}, and human-computer interaction~\cite{song2012continuous}. Nevertheless, due to large differences within the same class of actions, viewpoint variations, occlusions and changes in lighting conditions, action recognition still remains a challenging problem.

Consequently, there is a wide variety of action recognition approaches in the literature.  
One way to categorize them is based on the area features are computed on; \textit{global} approaches, where the entire image is used to generate features~\cite{weinland:inria-00544629,910878}, and \textit{local} approaches, where specific regions of interest are selected to generate features. One of the most popular approaches belonging to the second category is \emph{Dense Trajectories}
~\cite{wang:2011:inria-00583818:1}, in which every action is represented by a set of motion trajectories along which features are aligned and encoded using the Bag-of-Words (BoW) model~\cite{Li:2005:BHM:1068508.1069129}. 

Approaches based on Dense Trajectories are particularly effective when the amount of motion is high~\cite{Koperski14}. This is mainly because images in a video are densely sampled and tracked for generating the trajectories. However, Dense Trajectories, by definition, include trajectories of points that are irrelevant for action recognition due to background motion, noise, etc.; thus, resulting in the inclusion of irrelevant information. Furthermore, Dense Trajectories are typically generated using optical flow which fails to describe motion with radial orientation with respect to the image plane. 
Therefore, taking advantage of the availability of RGB-D cameras, we propose to redefine Dense Trajectories by giving them a local description power. This is achieved by clustering Dense Trajectories around human body joints provided by RGB-D sensors, which we will refer to as \emph{Localized Trajectories} henceforth. 

The proposed approach offers two main advantages. First, since we only consider trajectories that are localized around human body joints, our approach is more robust to large irrelevant motion estimates. As a consequence, actions which have similar motion patterns, but involving different body parts, are more easily distinguished. 
Second, our approach allows the description of the relationship of ``\textit{action-motion-joint}'', i.e.  an action is associated with both; a type of motion and joint location, in contrast to classical Dense Trajectories described by the relationship ``\emph{action-motion}'' where action is associated with a type of motion only. This is done by generating features around the Localized Trajectories based on the concept of local BoWs~\cite{lazebnik2006beyond}. One codebook is therefore constructed per group of Localized Trajectories. Each codebook corresponds to a specific body joint. 

For a better description of radial motion, we further propose to explore Localized Trajectories using the three modalities provided by RGB-D cameras.
Specifically, we introduce the \textit{3D Localized Trajectories} concept, which requires the estimation of scene flow, the displacement vector field in 3D, instead of optical flow.
Coupling 3D Trajectories and the corresponding motion descriptors with Localized Trajectories offers richer localized motion information, in both lateral and radial directions, allowing a better discrimination of actions. However, scene flow estimation is generally more noisy resulting in a less accurate temporal tracking of points. Thus, we propose to construct local codebooks by sampling trajectory-aligned features based on confidence and ambiguity metrics~\cite{Wang12}.

This paper is an extended version of~\cite{papadopoulos2017enhanced}. Compared to our previous work, the main contribution is the generalization of the proposed Localized Trajectories to 3D using RGB-D data. This extension is combined with a novel codebook construction scheme, suitable for tackling noisy feature samples. Moreover, an extensive comparison with state-of-the-art approaches is presented, along with evaluation on multiple datasets and novel discussions and analysis. 

In summary, the contributions of this paper are listed as follows:
\begin{enumerate}
\item A novel 2D Localized Trajectories concept is introduced, which utilizes body pose information in order to spatially group similar trajectories together.

    \item Localized Trajectories are extended from 2D to 3D thanks to the availability of depth data, which are directly used for 3D motion estimation.

   \item  
   A novel feature selection concept for a robust codebook construction is introduced. 
    \item An extensive experimental evaluation on several RGB-D datasets is presented to validate the discriminative power of the proposed approach.
\end{enumerate}

The remainder of the paper is organized as follows: in Section~\ref{sec:relatedwork}, a literature review of related works is given, followed by a detailed overview of background material in Section~\ref{sec:background}. The proposed approach is described in Section~\ref{sec:3dtrajectories} and Section~\ref{sec:3dtraj}. In Section~\ref{sec:experiments}, descriptions of different datasets, experimental setups, and results are presented. Finally, Section~\ref{sec:conclusion} concludes the paper and provides a perspective on future research directions.

\section{Related Work}
\label{sec:relatedwork}

In this section, we present some of the established action recognition approaches in the literature. First, we start by giving a general overview of RGB-D based action recognition approaches. Then, we focus on representations inspired by Dense Trajectories which are directly related to our work.

\subsection{Dense Trajectories Related Approaches}

\label{sec_rel_traj}

Initially introduced by Wang et al.  \cite{wang:2011:inria-00583818:1}, Dense Trajectories are classically generated by computing motion and texture features around motion trajectories. Due to their popularity, many researchers have extended this original formulation in order to enhance their performance \cite{wang2013action,Koperski14,wang2015action,jiang2012trajectory,ni2015motion}.

As a first attempt, Wang et al. \cite{wang2013action} proposed to reinforce Dense Trajectories by using the Random Sampling Consensus (RANSAC) algorithm to reduce the noise caused by motion. In addition to that, they have replaced the Bag-of-Visual-Words representation with Fisher Vectors.

Then, Koperski et al. \cite{Koperski14} suggested enriching motion trajectories using depth information. They proposed a model grouping the videos in two types: videos with high level of motion and others with low amount of motion. For the first group, an extension of Trajectory Shape Descriptor \cite{wang:2011:inria-00583818:1} which includes depth information has been used, while for the second group a novel descriptor called Speeded Up Robust Features (SURF) has been introduced in order generate local depth patterns.

To further improve the accuracy of recognition, Wang et al. \cite{wang2015action} proposed to use deep learned features instead of heuristic spatio-temporal local ones such as Trajectory-Shape Descriptor (TSD)~\cite{wang:2011:inria-00583818:1}, Histogram of Oriented Gradients (HOG)~\cite{dalal2005histograms}, Histogram of Optical Flow (HOF)~\cite{CRHV:CVPR09}, and Motion Boundary Histogram (MBH)~\cite{wang:2011:inria-00583818:1}. 

On the other hand, in \cite{jiang2012trajectory}, a novel approach to encode relations between motion trajectories has been presented. Global and local reference points have been used to compute Dense Trajectories, offering robustness to camera motion.

Finally, Ni et al. \cite{ni2015motion} had the idea of focusing on trajectory groups which contribute more importantly to a specific action by defining an optimization problem. Towards the same direction, Jhuang et al. \cite{jhuang2013towards} proposed the extraction of features around joint trajectories, increasing the discriminative power of the original Dense Trajectories approach \cite{wang:2011:inria-00583818:1}. 

Although all the aforementioned methods have shown their effectiveness, they unfortunately lack locality information related to the human body. This piece of information is crucial when actions include similar motion patterns performed by different body parts. For this reason, we propose a novel dense trajectory-based approach by taking into consideration the local spatial repartition of motion with respect of the human body.

\subsection{Action Recognition From RGB-D Data}

With the recent availability of affordable RGB-D cameras, a large effort in action recognition using both RGB and depth modalities has been made. For a more comprehensive state-of-the-art, we refer the reader to a recent survey~\cite{survey}, where RGB-D based action recognition methods have been grouped in two distinct categories (according to the nature of the descriptor), namely, \textit{learned representations}~\cite{deep1,deep2,deep3} and \textit{hand-crafted representations}~\cite{Wang12, oreifej2013hon4d, wang2012robust}. 
 Since this work bears interest to the description of actions using Dense Trajectories, we mainly focus on hand-crafted based approaches. In turn, they can be classified as follows: depth-based approaches, skeleton-based approaches and hybrid approaches.

The first class of methods extracts directly human motion information from depth maps~\cite{Xia,HOG3D,HOG2,foggia2013,shukla,oreifej2013hon4d, SNVpami,Slama,rahmani}.
The second group gathers approaches which make use of the 3D skeletons extracted from depth maps. During the past few years, a wide range of methods have been designed using this high-level modality \cite{zanfir,yang2012eigenjoints,LieGroup,devanne20153,amor2016action,Demisse_2018_CVPR_Workshops,ghorbel2018kinematic}.

Compared to depth-based descriptors, skeleton-based descriptors require low computational time, are easier to manipulate and can better discriminate local motions. However, they are more sensitive to noise since they widely depend on the quality of the skeleton. Thus, to reinforce action recognition, a third class of methods called \textit{hybrid} makes use of more than two modalities. These approaches usually exploit the skeleton information to compute local features using RGB and/or depth images. These local RGB-D based features have shown noteworthy potential~\cite{Wang12, wang2012robust,li2010action}.  
Inspired by this relevant concept which aims at computing local depth-based and RGB-based features around specific joints, we propose to adapt the same idea to Dense Trajectories which have been proven to be one of the most powerful action representations. 

\section{Background: Dense Trajectories for Action Recognition}
\label{sec:background}

Dense Trajectories have been initially introduced by Wang et al. \cite{wang:2011:inria-00583818:1}. They are constructed by densely tracking sampled points over an RGB video stream and constructing representative features around the detected trajectories. As mentioned in Section 1, Dense Trajectories have been proven to be very effective in action recognition. They owe mainly their success to the fact that they incorporate low-level motion information. Below, we overview the Dense Trajectories approach.

Let $\mathcal{V}$ be a sequence of \(N\) images. Subsequently, representative points are sampled from each image grid with a constant stepping size -- we denote each sampling grid position at frame $t$ as \(\mathtt{p}_t=(x_t,y_t)\). The point \(\mathtt{p}_{t}\) is then estimated in the next frame using a motion field \((u_t,v_t)\), derived by optical flow estimation \cite{farneback2003two} such that:
\begin{align}
\mathtt{p}_{t+1} = \mathtt{p}_{t} + \kappa * (u_t,v_t),
\label{eq:motion_est}
\end{align}
where \(\kappa\) is a median filter kernel at the position \(\mathtt{p}_{t+1}\). As a result, large motion changes between subsequent frames are smoothed. Furthermore, to avoid drifting,  trajectories longer than the assigned fixed length
are rejected. Applying~\eqref{eq:motion_est} on $L$ frames results a smoothed trajectory estimation of the point \(\mathtt{p}_t=(x_t,y_t)\). We denote the \(m^{th}\) dense trajectory as:
\begin{align}
\mathcal{P}^m = \{\mathtt{p}^m_{t_0},...,\mathtt{p}^m_{t_0+L}\},
\label{eq:set_traj}
\end{align}
\noindent with $\tau=[t_0, t_0+L] \subset [1,N]$, $m \in \{1,..,M\}$, \(t_{0}\) the first frame of the sequence $\mathcal{V}$ and $M$ the total number of generated trajectories.

The set of $M$ trajectories generated in~\eqref{eq:set_traj} is used to construct descriptors aligned along a spatio-temporal volume. In~\cite{wang:2011:inria-00583818:1}, four types of descriptors are used: TSD~\cite{wang:2011:inria-00583818:1}, HOG~\cite{dalal2005histograms}, HOF~\cite{CRHV:CVPR09}, and MBH~\cite{wang:2011:inria-00583818:1}. Each of the above descriptors is designed to capture distinctive spatio-temporal features of the occurring motion. As a final step, all of the descriptors are aggregated and encoded using BoWs -- one codebook of visual words per descriptor is constructed using K-means clustering so that the final features are represented by a unified histogram of word appearances. 

One of the main drawbacks of Dense Trajectories is that points on the image grid are sampled uniformly, which potentially leads to the inclusion of a significant amount of noise. Furthermore, the generated Dense Trajectories do no take into account the spatial human body structure. Thus, actions with similar motion patterns can potentially be confused during classification.

\begin{figure*}[!h]
\centering
\includegraphics[scale=0.35]{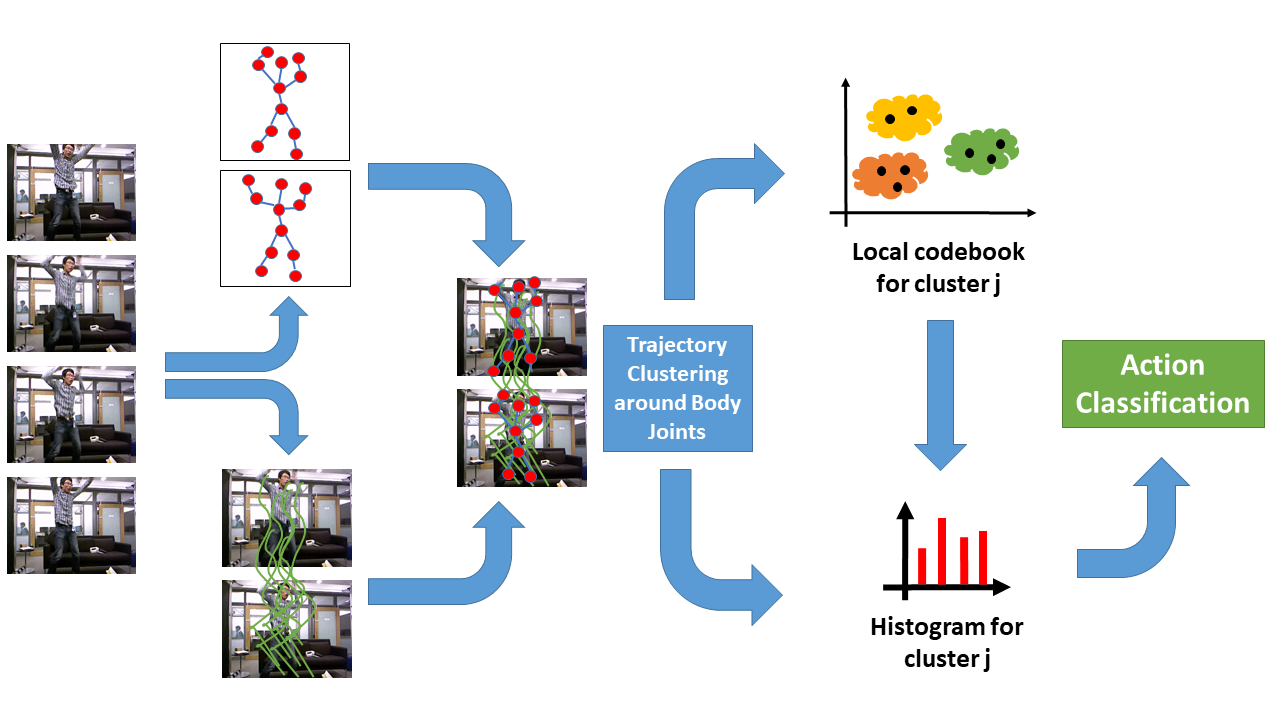}
\caption{Proposed 2D Localized Trajectories approach. From an RGB sequence, Dense Trajectories are generated and, then, clustered around body joints using RGB-D pose information (only 2D information is used). Finally, local codebooks, for every cluster $\mathsf{G}^j$, are constructed for the histogram representation of features. This feature representation is used in both training and testing phases of the classification.}
\label{fig_proposed}
\end{figure*}

\section{Localized Trajectories for Action Recognition}
\label{sec:3dtrajectories}

To enhance their robustness to irrelevant information, a reformulation of Dense Trajectories is proposed, called Localized Trajectories. The main idea of this new approach consists in attributing Dense Trajectories a local description in order to: 1) track the motion in specific and relevant spatial regions of the human body, more specifically around the joints. 2) remove redundant and irrelevant motion information, which can negatively affect the classifier performance.

To that end, the pose information through estimated 3D skeletons is used as prior information to estimate an optimal clustering configuration. \\
Let us consider the human skeleton extracted from RGB-D cameras composed of \(J\) joints and let us denote the trajectory of each skeleton joint $j$ as $\mathcal{Q}^j = \{\mathtt{q}^j_1, ..., \mathtt{q}^j_N\}$. 
Note that we assume that the joints are always well detected. We use the distance proposed by Raptis et al.~\cite{Raptis12} to group Dense Trajectories of an action around joints. Given a pair of dense and joint trajectories, respectively, $\mathcal{P}^m$ and $\mathcal{Q}^j$, which co-exist in the temporal range $\tau$, 
the spatio-temporal distance between two given trajectories is expressed using \eqref{eq_d_1_2} as follows:
\begin{equation}d({\mathcal{P}^m,\mathcal{Q}^j}) =\max_{t\in \tau} s_t \cdot \frac{1}{L} \sum_{t\in \tau} r_t,
\label{eq_d_1_2}
\end{equation}
such that \(s_t =||\mathtt{p}^m_{t}-\mathtt{q}^j_{t}||_2\) is the spatial distance and \(r_t = ||(\mathtt{p}^m_{t}-\mathtt{p}^m_{t-1})-(\mathtt{q}^j_{t}-\mathtt{q}^j_{t-1})||_2\) is the velocity difference between trajectories $\mathcal{P}^m$ and $\mathcal{Q}^j$. 
Then, an affinity matrix is computed between every pair of trajectories $(\mathcal{P}^m, \mathcal{Q}^j)$ using \eqref{eq_d_1_2} as:
\begin{equation}
b({\mathcal{P}^m,\mathcal{Q}^j}) = \exp(-d({\mathcal{P}^m,\mathcal{Q}^j})),
\label{affinity}
\end{equation}
where the measure $d(\mathcal{P}^m,\mathcal{Q}^j)$ penalizes trajectories with significant variation in spatial location and velocity. After a hierarchical clustering procedure which is based on the affinity score~\cite{Raptis12}, a membership indicator function specifies the cluster $\mathsf{G}^{j^*}$ of joint $j^*$ each trajectory belongs to.
\begin{align}
    \mathsf{G}^{j^*}=  \{ \mathcal{P}^m, \forall m \in \{1,..,M\} \text{ and }  \operatorname*{arg\,min}_{j\in J}b({\mathcal{P}^m,\mathcal{Q}^j}) &=  j^* \}.%
    \label{opt}
\end{align}
Furthermore, trajectories that are above a certain threshold of distance are rejected. 
This condition ensures that irrelevant and noise-resulting trajectories will not be considered, e.g, background motion.

\noindent \textbf{Feature Representation}: 
As discussed in~\cite{wang:2011:inria-00583818:1}, features can be computed along each trajectory and BoWs can be used to aggregate and encode the information. In such a case, however, a descriptor associated with each trajectory carries no locality information. On the contrary, we propose to exclusively assign trajectories and their corresponding descriptors to trajectory clusters. The main advantage of such a construction is that every trajectory-aligned descriptor does not only capture the spatio-temporal characteristics of the trajectory but it carries its location as well. Thus, we construct a local codebook for each trajectory group $\mathsf{G}^j$. During feature encoding, one histogram is constructed per joint cluster and per descriptor denoted by ${H}^j$:
\begin{align}
{H}^j =\Big[{H}_{TSD}^j\big|{H}_{HOG}^j\big|{H}_{HOF}^j\big|{H}_{MBH}^j \Big].
\label{small_hist}
\end{align}
The subscripts of the individual histograms identify the type of descriptors. Finally, an action video is represented by the concatenation of the individual joint histograms in a final histogram ${H}$, as follows:
\begin{align} 
{H} = \bigcup_{j=1}^{J} {H}^{j}.
\label{big_hist}
\end{align}
The general overview of our approach is illustrated in Fig.~\ref{fig_proposed}.

\section{3D Trajectories and Aligned Descriptors}
\label{sec:3dtraj}

\begin{figure*}[!t]
\centering
\begin{tabular}[b]{c} 
\includegraphics[scale = 0.56]{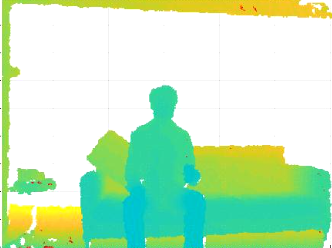}\\
(a)
\label{fig_first_case} 
\end{tabular}
\begin{tabular}[b]{c}
\includegraphics[scale = 0.56]{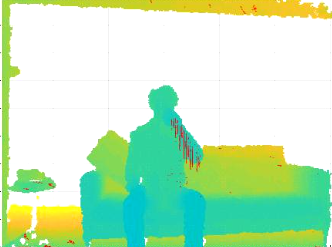}\\
(b)
\label{fig_second_case}
\end{tabular}
\begin{tabular}[b]{c}
\includegraphics[scale = 0.56]{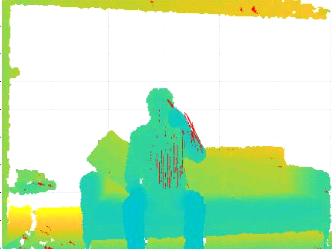}\\
(c)
\label{fig_third_case}
\end{tabular}
\\
\begin{tabular}[b]{c}
\includegraphics[scale = 0.58]{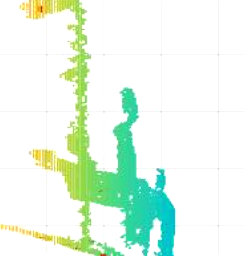}\\
(d)
\label{fig_forth_case}
\end{tabular} \hspace{0.6cm}
\begin{tabular}[b]{c}
\includegraphics[scale = 0.58]{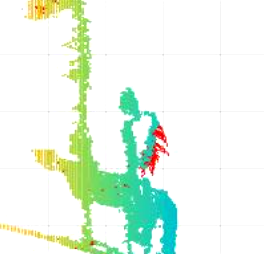}\\
(e)
\label{fig_fifth_case}
\end{tabular} \hspace{0.6cm}
\begin{tabular}[b]{c}
\includegraphics[scale = 0.58]{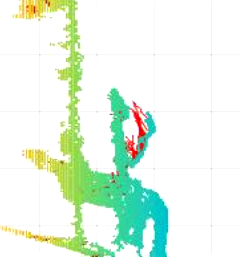}\\
(f)
\label{fig_sixth_case}
\end{tabular}
\caption{Scene flow-generated motion trajectories. Three phases of the same action are illustrated: In (a), (b) and (c) the frontal view of a subject drinking water is displayed as a point cloud, along with the corresponding motion trajectories in red. The same sequence is illustrated from the side in (d), (e) and (f). The capture of both lateral and radial motion shape is clearly depicted.}
\label{fig_3dt}
\end{figure*}

Dense Trajectories, generated via optical flow, offer adequate performance when used for tracking movements that are lateral to the image plane. However, they struggle to track motion that happens radially, due to the fact that the occurring motion is subtle with respect to the 2D image plane. Consequently, in this subsection, we propose to extend localized Dense Trajectories to RGB-D input video stream by replacing optical flow with scene flow. The generated 3D trajectories are suitable for tracking motion in both lateral and radial directions 
as illustrated in Fig.~\ref{fig_3dt}.

\subsection{Scene Flow Estimation Using RGB-D Data}
\label{sec:sceneflow}

To generalize the concept of Dense Trajectories from 2D to 3D, we propose to make use of the 3D extension of optical flow, called scene flow. Thanks to the emergence of RGB-D cameras, numerous approaches have been proposed to estimate scene flow from depth maps, e.g. the \emph{Primal-Dual Framework for Real-Time Dense RGB-D Scene Flow} (PD-Flow) algorithm~\cite{jaimez2015primal}, the \textit{Dense semi-rigid scene flow estimation} \cite{quiroga2014dense} and the \textit{Layered RGBD scene flow estimation} \cite{sun2015layered}. 

The scene flow $\mathbf{\Omega}$ is linearly dependent on the depth motion field  $\mathbf{S}=(u, v, w)$, where $w$ is the range flow. It is computed by mapping $\mathbf{S}$ to the 3D world coordinate system as below:

\begin{equation}
\mathbf{\Omega} = \begin{pmatrix} 
\frac{Z}{f_x} & 0 & \frac{X}{Z}\\
0 & \frac{Z}{f_y} & \frac{Y}{Z}\\
0& 0 & 1
\end{pmatrix} 
\mathbf{S}^{T},
\label{eq:proj}
\end{equation}
where $f_x$ and $f_y$ are the camera focal lengths, and $X,Y,Z$ are the 3D world coordinates of a specific point. On the other hand, The depth motion fields are estimated as a solution of a global variational problem, defined as:
\begin{align}
\label{minim}
\min_\mathbf{S} \{E_D(\mathbf{S}) + E_R(\mathbf{S})\},
\end{align}

\noindent where \(E_D(\mathbf{S})\) is a data term defined as the combined measure of the photometric and geometric inconsistency of successive depth and intensity images and \(E_R(\mathbf{S})\) is defined as a regularizer term.

We choose PD-Flow \cite{jaimez2015primal} to estimate a dense scene flow field from an RGB-D video stream, since it has been shown to be one of the fastest and most accurate algorithms.

\subsection{3D Localized Trajectories}

To estimate the 3D trajectories using the scene flow, we start by uniformly sampling points from the 2D image grid. In this context, we define pixel coordinates as $(x,y)$. Similar to~\cite{wang:2011:inria-00583818:1}, we reject points belonging to homogeneous areas. Next, each of the sampled points are 
mapped to a standard 3D world coordinate system using the inverse of the intrinsic camera parameter matrix as described below:
\begin{align}
\begin{pmatrix}
X\\
Y\\
Z
\end{pmatrix} =
\begin{pmatrix}
\dfrac{(x-c_x)D}{f_x} &
\dfrac{(y-c_y)D}{f_y} &
D
\end{pmatrix}^T,
\label{eq:pro_scene}
\end{align}
where $c_x$, $c_y$ are the image plane central point coordinates, $f_x$ and $f_y$ are the respective x and y components of the focal length and $D$ is the depth value. Subsequently, trajectories of the mapped 3D points are estimated using~\eqref{eq:motion_est}, except that the motion field is now based on an estimated scene flow. The estimated \textit{3D Dense Trajectories} are denoted as:
\begin{align}
 (X_{t+1}, Y_{t+1}, Z_{t+1}) = (X_{t}, Y_{t}, Z_{t}) + \mathbf{\Omega_{t}} ,  
 \label{eq_3D}
\end{align}
where $\mathbf{\Omega_{i}}$ is the scene flow field. Correspondence between estimated 3D points, with scene flow, and image pixels is derived by solving \eqref{eq:pro_scene} in terms of $(x,  y,  D)^T$.

The above procedure is repeated recurrently until each of the 3D trajectories reach the fixed temporal length we have set. Similar to~\cite{wang:2011:inria-00583818:1}, trajectories with sudden displacements or small overall spatial length are considered irrelevant and are removed.

In depth maps, texture information is not present. Thus, in our case, only motion descriptors are considered. Three types of descriptors are used: \textit{3D Trajectory Shape Descriptor (3DTSD), Histogram of Scene Flow \cite{holte2012local} (HSF)}, and \textit{3D Motion Boundary Histogram (3DMBH)}. 3DTSD is based on the original idea of the TSD for Dense Trajectories \cite{wang:2011:inria-00583818:1}. For each trajectory, the normalized displacement vector is computed. The HSF descriptor captures the orientation and the magnitude of the local scene flow field. For a spatio-temporal volume aligned around a 3D trajectory, the orientation of the 3D displacement is calculated using the azimuth $\theta_{x,y}$ and elevation  $\theta_{y,z}$ angles  formed by consecutive points as:

\begin{equation}
\theta_{xy} = \frac{\Delta Y_t}{\Delta X_t} \hspace{0.4cm} \text{and} \hspace{0.4cm}
\theta_{yz} = \frac{\Delta Z_t}{\Delta Y_t}.
\label{angles}
\end{equation}
For the histogram construction, the 4D space is quantized into a fixed number of bins. Similarly, the 3DMBH is based on the same idea as HSF. First, the derivative of the scene flow field is computed and, then, for every pair of coordinates, the orientation angle is estimated.

3D Trajectories are adapted to \textit{3D Localized Trajectories} by following the procedure described in Section~\ref{sec:3dtrajectories}. Similarly as before, we propose to enhance the discriminative power of 3D Trajectories by grouping them around 3D body joints. Hence, \eqref{eq_d_1_2}, \eqref{affinity} and \eqref{opt} are adapted accordingly to incorporate all three dimensions of 3D trajectories $\mathcal{P}_{3D}^m$ and 3D joint trajectories $\mathcal{Q}_{3D}^j$. Then, during feature encoding, every histogram of joint clusters $\mathsf{G}^j$ defined in \eqref{small_hist} is modified to include the descriptors used in this context, becoming:
\begin{align}
{H}^j =\Big[{H}_{3DTSD}^j\big|{H}_{HSF}^j\big|{H}_{3DMBH}^j \Big].
\label{small_hist_3d}
\end{align}

\subsection{Feature Selection for Codebook Construction}

While 3D Trajectories are advantageous in capturing radial motion, they are notably more noisy compared to Dense Trajectories, due to the scene flow estimation. 
As a result, the quality of the codebooks is degraded, unfavorably affecting the general performance of the proposed approach. In order to enhance it, we propose to select features according to the classifier \textit{confidence} and \textit{ambiguity} probabilistic metrics. Confidence is the classifier ability to quantify its predictions reliability, while ambiguity indicates the number of classes the classifier outputs for every prediction.  
The goal is to encode trajectory features $F^m$ using codebooks constructed by sampling features from a training set $M_r$, 
which maximize the classifier confidence $\mathcal{C}$ and minimize ambiguity $\mathcal{A}$ metrics. 
The confidence $\mathcal{C}$ and ambiguity $\mathcal{A}$ metrics are defined as:
\begin{align}
\mathcal{C} &= \underset{m\in M_r}{\mathrm{median}} (log(Pr(l_m=a|F^m))),
\end{align}
\begin{center}\text{and} \end{center}
\begin{align}
\mathcal{A} &= \sum_{m\not\in M_r}(log(Pr(l_m=a|F^m))),
\label{metrics}
\end{align}
\noindent 
where $Pr(l_m=a|F^m)$ is the posterior probability of label $a$ given feature $F^{m}$.

\section{Experimental Evaluation}
\label{sec:experiments}

In this section, we evaluate the proposed approaches on $5$ challenging datasets: MSRDailyActivity3D \cite{Wang12}, Online RGB-D Action (ORGBD) \cite{yu2014discriminative}, G3D Gaming Action \cite{bloom2016hierarchical}, Watch-n-Patch \cite{Wu_2015_CVPR} and KARD datasets \cite{gaglio2015human}. First, a brief description of each dataset is given 
followed by description of the experimental setups. 
Then, the obtained results are reported and extensively analyzed.
\\
\\
\noindent \textbf{Datasets and Experimental Settings}: The first dataset used for the experimental evaluation is the MSRDailyActivity 3D \cite{Wang12}. In this dataset, $10$ actors perform $16$ daily activities, which in some cases involve human-object interaction. 
The dataset is captured by the Kinect v1 device, providing therefore RGB, depth and skeleton modalities. A distinctive characteristic of this dataset is that every actor repeats each action twice in both sitting and standing position. For the experiments, we follow a cross-splitting protocol as in \cite{Wang12}, where half of the subjects are used for training and the rest for testing. 

The second dataset is called Online RGB-D Action (ORGBD) \cite{yu2014discriminative}. It can be used for both action recognition and action detection and includes $7$ common types of human-object interaction related to the living room environment. 
Three sets of video sequences are collected using a Kinect sensor. Thus, RGB, depth and skeleton modalities are available. The first set is captured in the context of action recognition in the Same Environment, whereas the second set is acquired for cross-environment action recognition and the third for on-line action detection. The splitting protocol requires two fold cross-validation for the same-environment scenario, whereas, for cross-environment action recognition, training and testing sets should include different environments \cite{yu2014discriminative}. 

One challenging dataset used for the evaluation is the G3D Gaming Action Dataset \cite{bloom2016hierarchical}. This Kinect-acquired dataset can be used for both action recognition and temporal action detection. It consists of $10$ subjects performing $20$ gaming actions 
which are grouped into $7$ gaming scenarios, which are: \textit{Fighting, playing golf, playing tennis, bowling, first person shooter, driving a car} and \textit{miscellaneous}. The first $5$ actors are used for training and the rest are used for testing \cite{bloom2016hierarchical}.

Watch-n-Patch \cite{Wu_2015_CVPR} dataset, which was introduced by the Cornell University, is also utilized. This dataset includes $21$ types of actions ($10$ in an office and $11$ in a kitchen) which involve interactions with $23$ types of objects. $7$ subjects perform $2$-$7$ actions in every of the $458$ videos. The dataset was recorded using a Kinect v2 camera. This dataset distinguishes itself by a high intra-class variability since the subjects perform different combinations of actions and order them differently each time. 
For the experiments, we use the provided splitting protocol proposed in \cite{Wu_2015_CVPR}, where, for every environment, almost half of the videos are used for training and the rest for testing.

The last dataset used for evaluation is called Kinect Activity Recognition Dataset (KARD) \cite{gaglio2015human}. It contains $18$ action classes 
which are performed by $10$ subjects ($9$ males and $1$ female) where half of them are used for training and the other half for testing, as proposed in \cite{gaglio2015human}. The dataset was captured by a Kinect device and consequently contains the three RGB-D modalities: RGB images, depth maps and 3D skeletons.
\\
\\
\noindent \textbf{Implementation Details}: For extracting Dense Trajectories and features from videos, we use the implementation provided by the authors in \cite{wang:2011:inria-00583818:1}\footnote{\url{https://lear.inrialpes.fr/people/wang/dense_trajectories}}. The trajectory temporal length is fixed to $15$ frames. The features are computed on a spatio-temporal volume of $32\times 32 \times15$ aligned on the trajectory, as suggested in \cite{wang:2011:inria-00583818:1}. This volume is further divided into $2\times 2 \times 3$ cells, where the histograms of the descriptors are computed. In the case of 3D trajectories, we use the same parameters for the spatio-temporal volume. The number of histogram bins for the 2D trajectories is set to $8$ for HOG and MBH descriptors and $9$ for HOF descriptor, whereas for 3D trajectories case we use $9$-bin histograms for every descriptor. The distance threshold for each trajectory is set to $0.02$. Moreover, a linear SVM is employed for classification.

For each one of the aforementioned datasets, we report the obtained recognition accuracy using the proposed Localized Trajectories and compare it to the classical Dense Trajectories and recent state-of-the-art approaches.
In the following, we denote the original dense trajectory approach \cite{wang:2011:inria-00583818:1} by \textbf{Dense Trajectories}. We refer to the 2D proposed approach as \textbf{2D Localized Trajectories}. Similarly, the proposed 3D extension of the classical and the local Dense Trajectories are respectively called \textbf{3D Dense Trajectories} and \textbf{3D Localized Trajectories}.

The number of skeleton joints defines the number of clusters. Subsequently, in MSRDailyActivity3D, ORGBD and G3D datasets, the skeletons are composed of $20$ joints, while in Watch-n-Patch and KARD datasets, they are respectively formed by $25$ and $15$ joints. We, also, choose empirically $2000$ trajectories per video in order to construct the codebooks and $128$ words per cluster and per descriptor for every dataset. 

\begin{figure}[ht!]
\centering
\begin{tabular}[b]{c}
\includegraphics[width=0.65\textwidth]{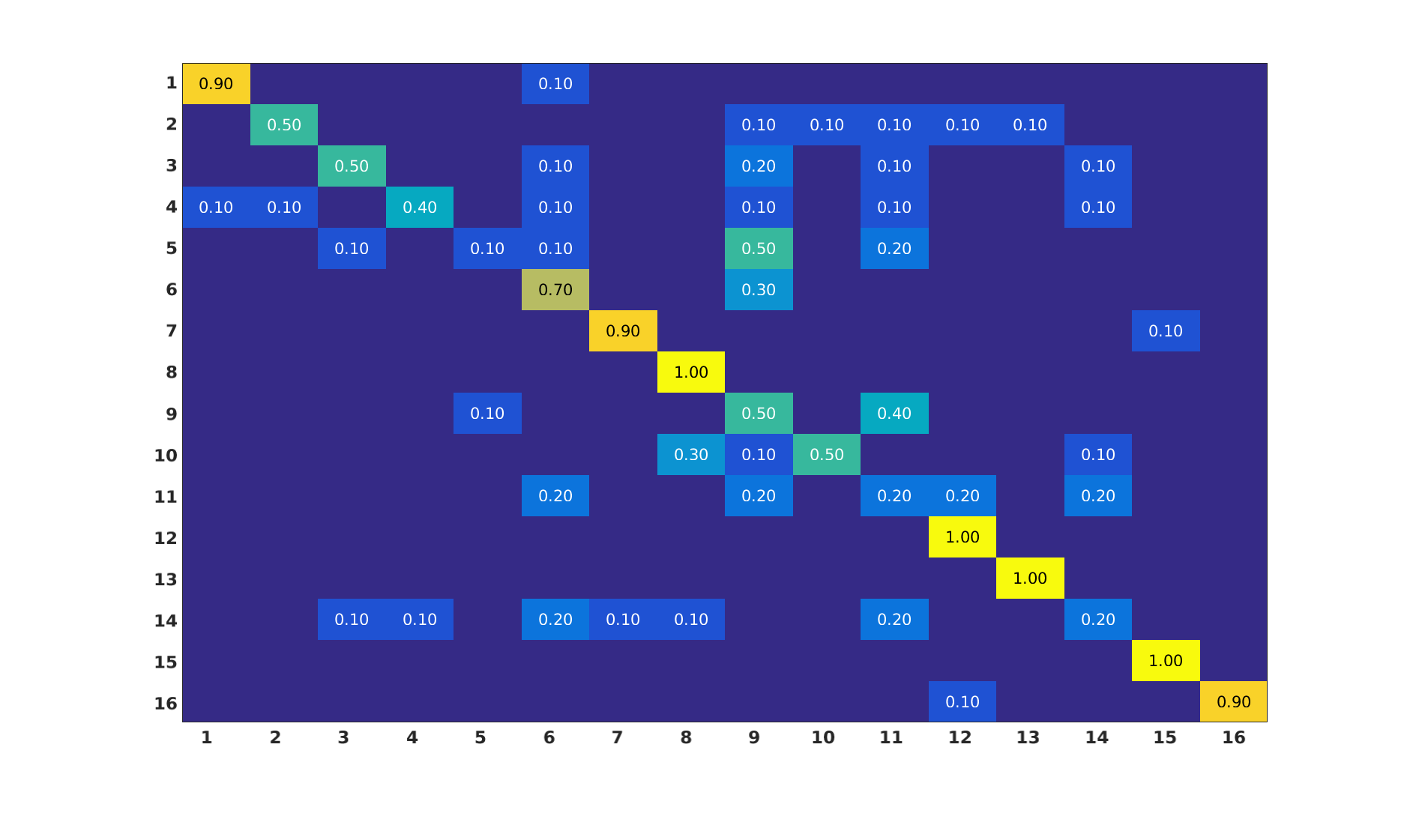}\\
(a)
\label{fig_first_case_msr}
\end{tabular}
\begin{tabular}[b]{c}
\includegraphics[width=0.65\textwidth]{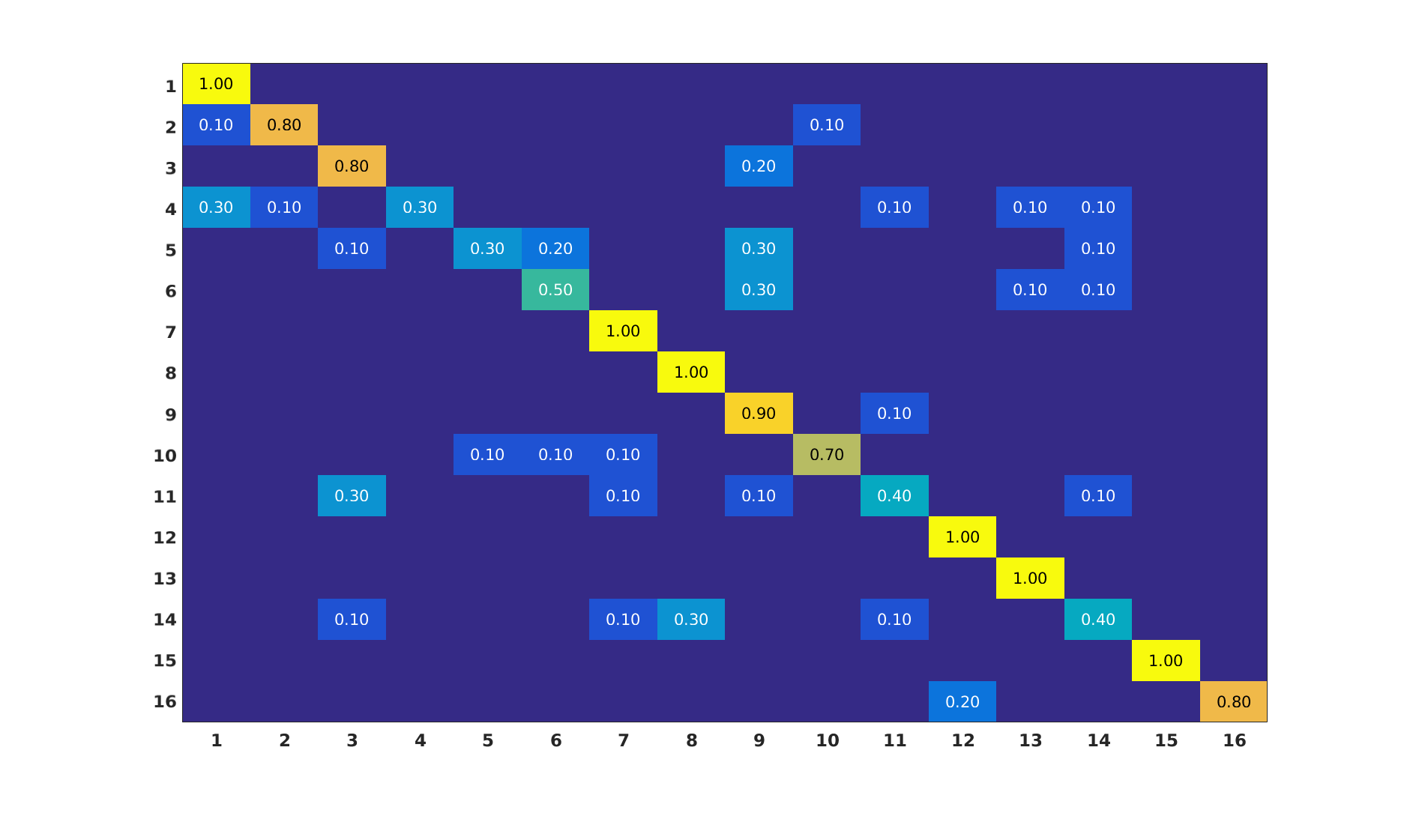}\\
(b)
\label{fig_second_case_msr}
\end{tabular}
\begin{tabular}[b]{c}
\includegraphics[width=0.65\textwidth]{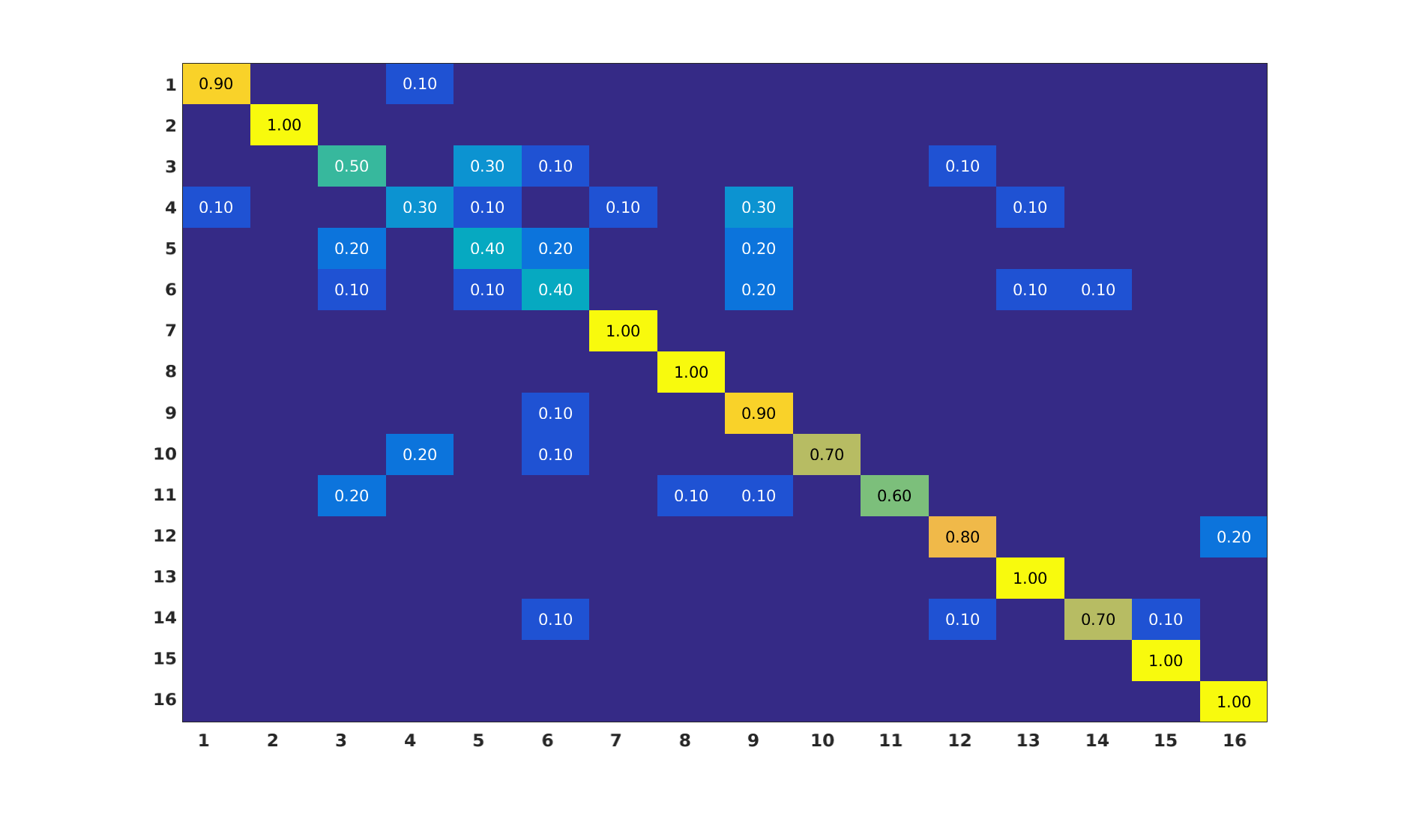}\\
(c)
\label{fig_third_case_msr}
\end{tabular}
\caption{Confusion matrices obtained for Dense Trajectories, 2D Localized Trajectories and 3D Localized Trajectories approaches on MSR DailyActivity 3D dataset. Actions list: \textit{1) Drink 2) Eat 3) Read book 4) Call cellphone 5) Write on a paper 6) Use laptop 7) Use vacuum cleaner 8) Cheer up 9) Sit still 10) Toss paper 11) Play game 12) Lie down on a sofa 13) Walk 14) Play guitar 15) Stand up 16) Sit down}.}
\label{fig_msrconf}
\end{figure}

\begin{table}[h!]
\renewcommand{\arraystretch}{1.3}
\caption{Mean accuracy of recognition (\%) on MSR DailyActivity 3D dataset for Dense Trajectories and 2D Localized Trajectories approaches against literature.}
\label{table_msr}
\centering
\begin{tabular}{|c||c|}
\hline
Method & Mean accuracy\\
\hline
\hline
Dynamic Temporal Warping \cite{muller2006motion} & 54.0\%\\
\hline
Local HON4D \cite{oreifej2013hon4d} & 80.0\%\\
\hline
Moving Pose \cite{zanfir2013moving} & 73.8\%\\
\hline
3D Trajectories \cite{Koperski14} & 72.0\%\\
\hline
Skeleton only \cite{Wang12} & 68.0\%\\
\hline
Skeleton \& LoP \cite{Wang12} & 85.8\%\\
\hline
Naive-Bayes-NN \cite{yang2012eigenjoints} & 73.8\%\\
\hline
Dense Trajectories \cite{wang:2011:inria-00583818:1} & 64.4\%\\
\hline
3D Dense Trajectories & 48.8\%\\
\hline
\textbf{2D Localized Trajectories} & \textbf{74.4\%}\\
\hline
\textbf{3D Localized Trajectories} & \textbf{76.3\%}\\
\hline
\end{tabular}
\end{table}

\begin{figure*}[h!]
\centering
\begin{tabular}[b]{c}
\includegraphics[scale = 0.185]{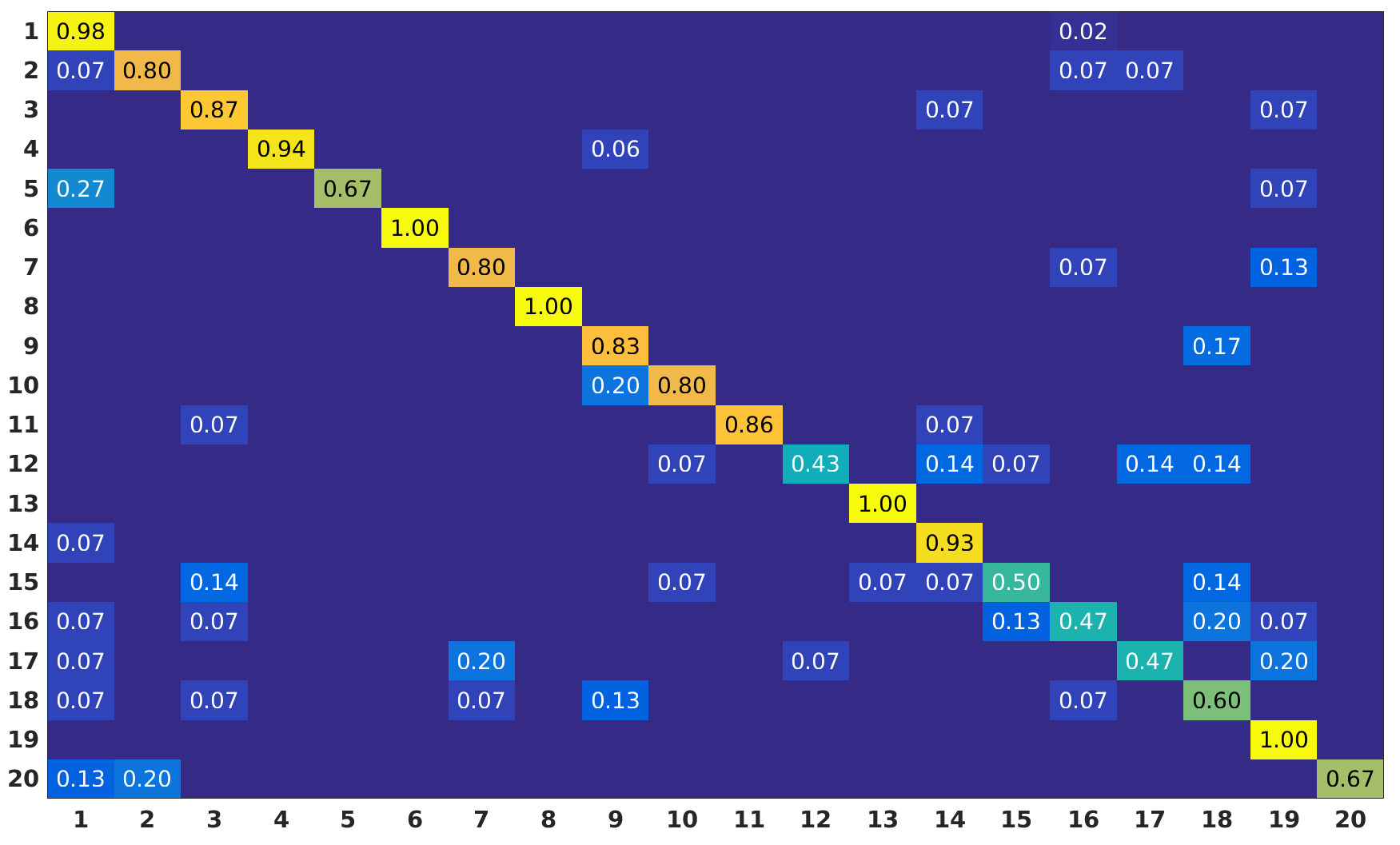} \\
(a)
\label{fig_first_case_g3d}
\end{tabular}
\begin{tabular}[b]{c}
\includegraphics[scale = 0.185]{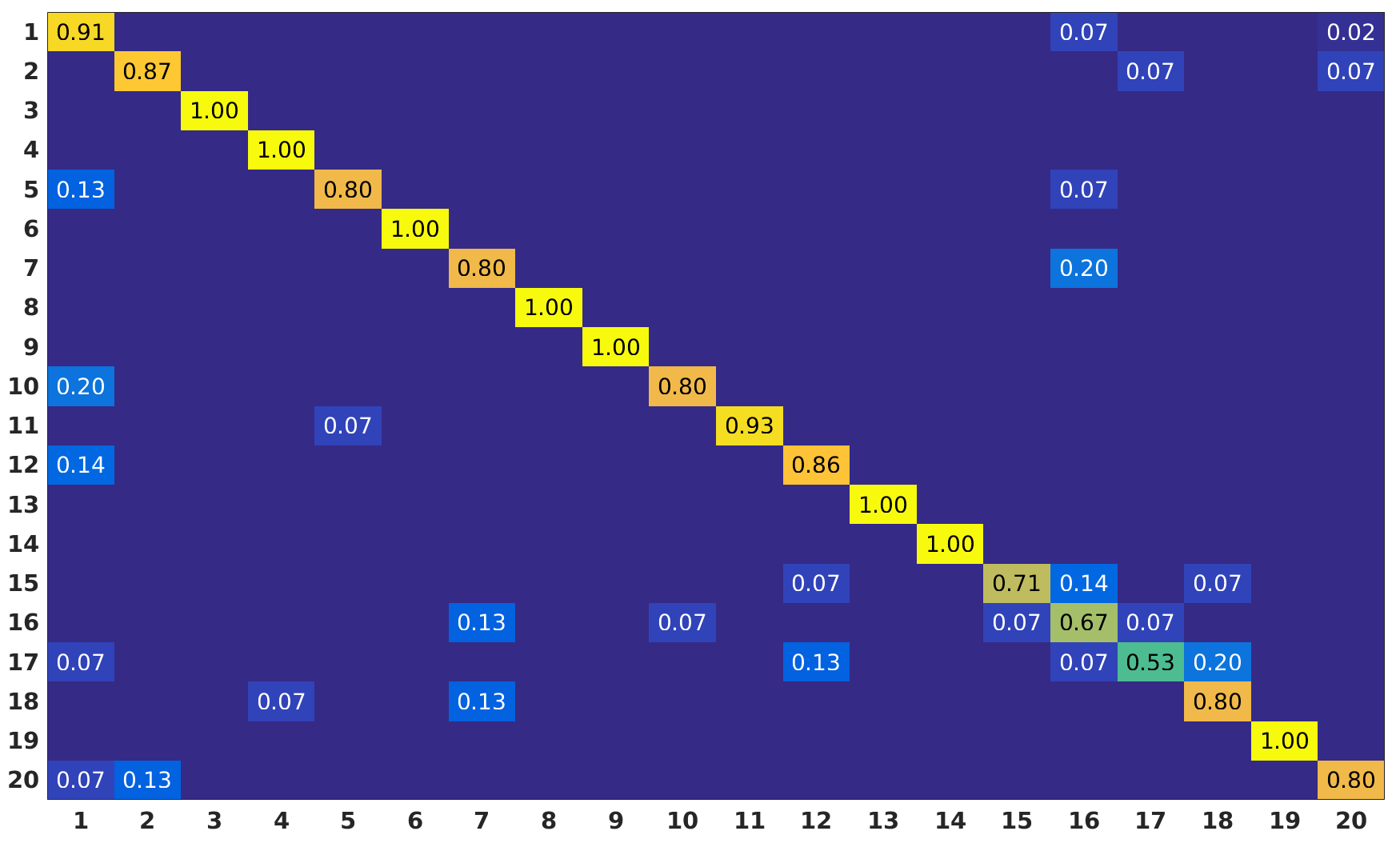} \\
(b)
\label{fig_second_case_g3d}
\end{tabular}
\caption{Confusion matrices obtained for Dense Trajectories (a) and 2D Localized Trajectories (b) approaches on G3D dataset. Actions list: \textit{1) Aim \& Fire Gun 2) Clap 3) Climb 4) Crouch 5) Defend 6) Flap 7) Golf Swing 8) Jump 9) Kick Left 10) Kick Right 11) Punch Left 12) Punch Right 13) Run 14) Steer 15) Tennis Serve 16) Tennis Swing Backhand 17) Tennis Swing Forehand 18) Throw Bowling Ball 19) Walk 20) Wave}}
\label{fig_g3dconf}
\end{figure*}

\begin{table}[h!]
\renewcommand{\arraystretch}{1.3}
\caption{Mean accuracy of recognition (\%) on G3D dataset for Dense Trajectories and 2D Localized Trajectories approaches against literature.}
\label{table_g3d}
\centering
\begin{tabular}{|c||c|}
\hline
Method & Mean accuracy\\
\hline
\hline
Dynamic Time Wrapping \cite{leightley2014exemplar} & 86.3\%\\
\hline
Weighted Graph Matching \cite{xiao2015motion} & 89.2\%\\
\hline
Adaptive Graph Kernels \cite{li20163d} & 84.8\%\\
\hline
Histogram \cite{barnachon2014ongoing} & 79.5\%\\
\hline
LPP \& BoW \cite{fotiadou2014activity} & 87.5\%\\
\hline
Spatial Graph Kernels \cite{kishore2018spatial} & 95.7\%\\
\hline
Dense Trajectories \cite{wang:2011:inria-00583818:1} & 80.1\%\\
\hline
\textbf{2D Localized Trajectories} & \textbf{87.8\%}\\
\hline
\end{tabular}
\end{table}

\subsection{Performance of 2D Localized Dense Trajectories}

In this subsection, an analysis of the obtained results is provided. First, we compare the performance of our approach against Dense Trajectories and other state-of-the-art methods. 
Later, we discuss some of the limitation of 2D Localized Trajectories.

\subsubsection{2D Localized Dense Trajectories vs Dense Trajectories}

Since the aim of this work is to improve the discriminative power of classical Dense trajectories, we start by comparing our proposed 2D Localized Dense Trajectories with them. 
The results obtained on the five benchmarks prove the superiority of the proposed 2D Localized Trajectories. As reported in Table~\ref{table_msr}, Table~\ref{table_g3d}, Table~\ref{table_orgbd_se}, Table~\ref{table_wnp} and Table~\ref{table_kard}, 2D Localized Dense Trajectories improve the accuracy by $10\%$, $7.7\%$, $3.1\%$, $16\%$, $13.8\%$ and $0.4\%$ on MSRDailyAvtivity3D, G3D, ORGB (same-environment settings), ORGB (cross-environment settings), Watch-n-Patch and KARD, respectively, compared to the classical Dense Trajectories \cite{wang:2011:inria-00583818:1}.

The reported results reflect the ability of 2D Localized Trajectories to distinguish actions with similar motion patterns that are performed by different body parts. This is shown in various cases when comparing confusion matrices obtained for 2D Localized Trajectories and Dense Trajectories. For instance, in the confusion matrices of G3D dataset in Fig.~\ref{fig_g3dconf}, 2D Localized Trajectories boost the performance of the following action pairs: \textit{Punch Right}-\textit{Punch Left} and \textit{Kick Right}-\textit{Kick Left}. Also, in the same dataset, the recognition accuracy of both \textit{Tennis Swing Backhand} and \textit{Throwing Bowling Ball} activities which include similar motion shapes is improved by $20\%$ and $6\%$, respectively. Furthermore, the accuracy of \textit{Drinking} and \textit{Reading Book} classes in ORGBD dataset  
is increased by $33\%$ and $31\%$, respectively (see Fig.~\ref{fig_orgbdconf}).

Another example of this enhancement can be the pair of actions \textit{Defend} and \textit{Aim \& Fire Gun} in G3D dataset. 
The motion shapes of both action classes are similar, since both of them include arm raising. Nevertheless, the first is performed using both arms and the second by using only one arm. As we can see in Fig.~\ref{fig_g3dconf}, the performance obtained for the action \textit{Defend} is improved by $13\%$ and the confusion with the action \textit{Aim \& Fire Gun} is reduced by $14\%$. In addition, in the same dataset, actions \textit{Wave} and \textit{Clap} have similar lateral motion and using the classical Dense Trajectories made their distinction challenging. However, with the use of 2D Localized Trajectories, motion trajectories were assigned to only one hand cluster in \textit{Wave} action and to both hands in \textit{Clap} action, reducing the confusion between these classes. This results in an accuracy boost of $13\%$ in \textit{Wave} class, as it is shown in Fig.~\ref{fig_g3dconf}.

Moreover, in scenarios with full-body motion, such as the kitchen environment in Watch-n-Patch dataset, 2D Localized Trajectories outperform the Dense Trajectories approach as shown in Fig.~\ref{fig_wnpconf}. Clusters isolate specific motion of body parts, therefore motion patterns related to the action can be identified more effectively. 

\subsubsection{Comparison with 3D-Based State-of-the-Art Approaches}

Our 2D Localized Trajectories approach has shown competitive performance compared to 3D-based state-of-the-art approaches. 
In ORGBD dataset, we achieve the second best performance in Same Environment setting (Table~\ref{table_orgbd_se}). We manage to match the state-of-the-art results of \cite{Wang12} in Cross Environment settings and, at the same time, increase the mean accuracy by $16\%$ over the Dense Trajectories. 

\begin{figure*}[h!]
\centering
\begin{tabular}[b]{c}
\includegraphics[width=0.77\textwidth]{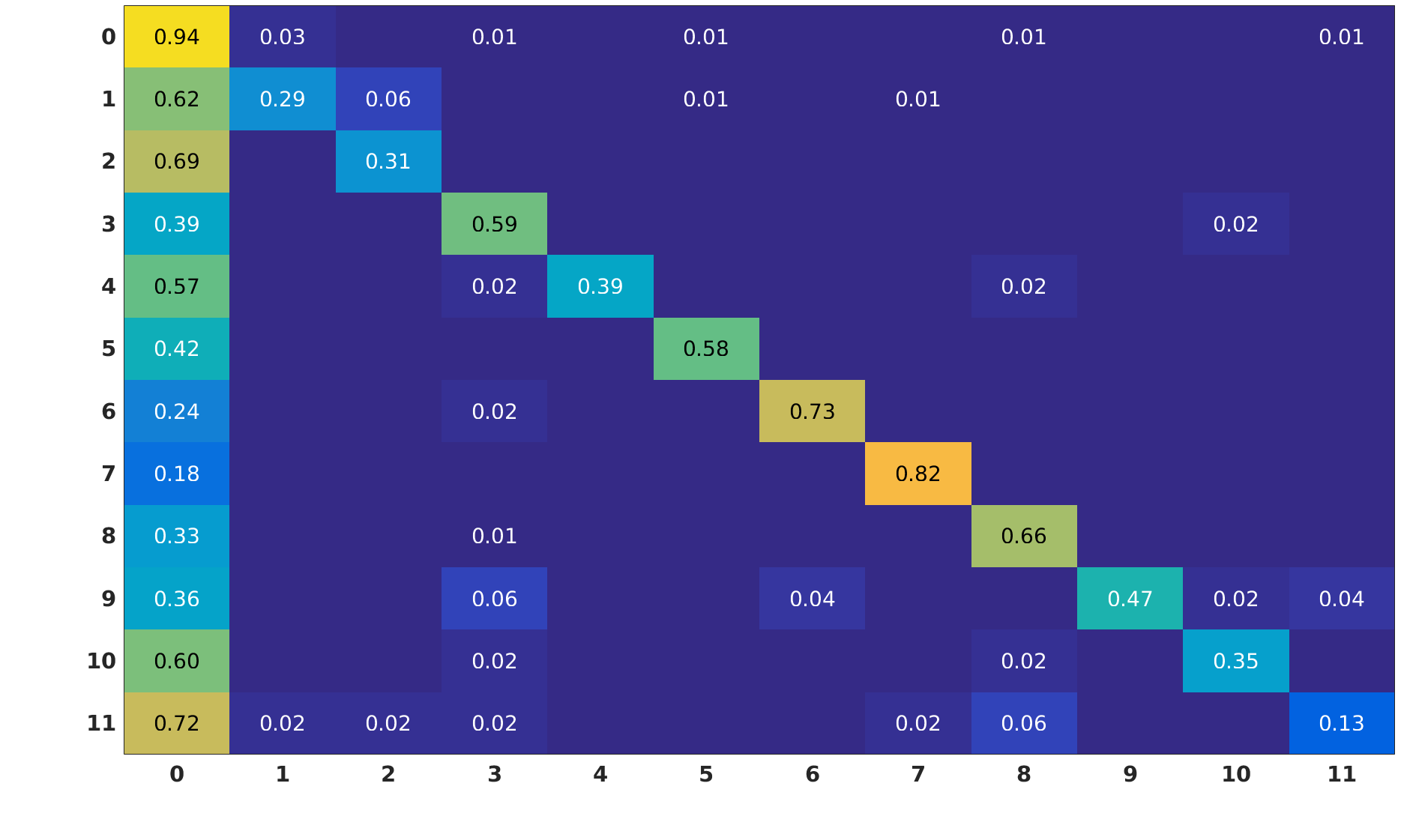}\\
(a)
\label{fig_first_case_wnp}
\end{tabular}
\begin{tabular}[b]{c}
\includegraphics[width=0.77\textwidth]{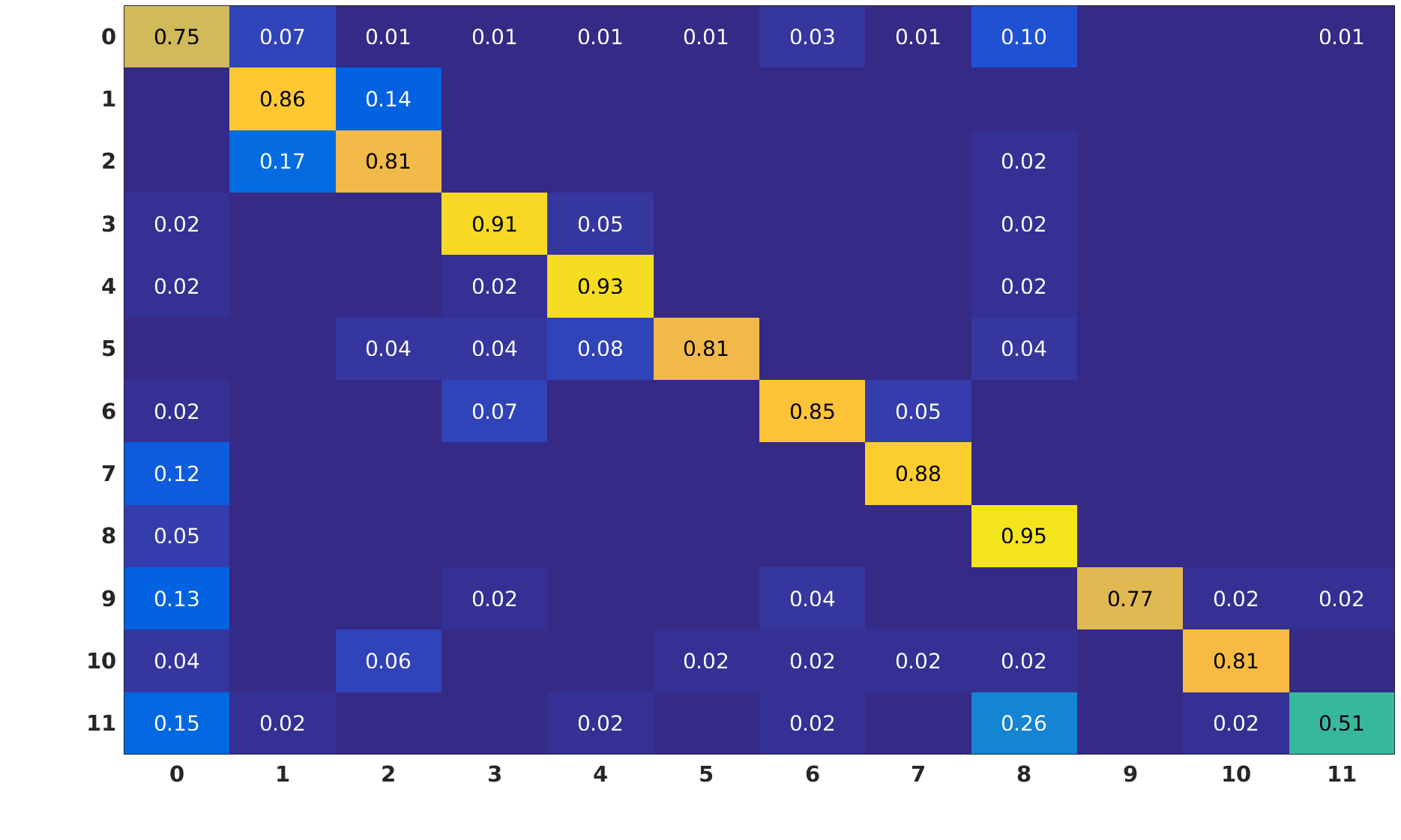}\\
(b)
\label{fig_second_case_wnp}
\end{tabular}
\caption{Confusion matrices obtained for Dense Trajectories (a) and 2D Localized Trajectories (b) approaches (Watch-n-Patch) in the kitchen environment. The action labels are: \textit{0) no-action, 1) fetch-from-fridge, 2) put-back-to-fridge, 3) prepare-food, 4) microwaving, 5) fetch-from-oven, 6) pouring, 7) drinking, 8) leave-kitchen, 9) fill-kettle, 10) plug-in-kettle, 11) move-kettle}.}
\label{fig_wnpconf}
\end{figure*}

In Watch-n-Patch dataset, 
the 2D Localized Trajectories improved the performance of the Dense Trajectories by $2.3\%$ in the office environment and by $25.3\%$ in the kitchen environment, as illustrated in Table~\ref{table_wnp}. The discriminative power of our approach boosts the performance of every action class, especially in the kitchen environment, as it can be observed in Fig.~\ref{fig_wnpconf}. On this dataset, we compare our work only with Dense Trajectories. 
To the best of our knowledge, there is no work in the literature reporting offline action recognition accuracy on it, since this dataset has been initially acquired for action detection.

In G3D dataset case, the results in Table~\ref{table_g3d} indicate that the 2D Localized Trajectories approach performes adequately enough compared to state-of-the-art 3D concepts, despite the fact that it includes a significant amount of radial motion. The obtained results in Table~\ref{table_g3d} show that our method was the third best performing, without utilizing depth or 3D skeleton modalities.

In KARD dataset, our approach based on the 2D Localized Trajectories outperforms almost all state-of-the-art approaches, with a score of $98.2\%$, except JTMI \& LBP \& FLD \cite{ahmed2016joint} which reaches a slightly superior score with only $0.3\%$ of difference.

The 2D Localized Trajectories approach offers the second largest improvement on MSRDailyActivity3D dataset, by $10\%$ compared to Dense Trajectories as it is depicted in Table~\ref{table_msr}. Apart from that, its performance was slightly inferior to the performance other state-of-the-art approaches, since it came third in average accuracy, behind Local HON4D \cite{oreifej2013hon4d} and Skeleton \& LoP \cite{Wang12}.

\subsubsection{Limitations of 2D Localized Dense Trajectories }

\begin{figure}[t!]
\centering
\begin{tabular}[b]{c}
\includegraphics[width=0.8\textwidth]{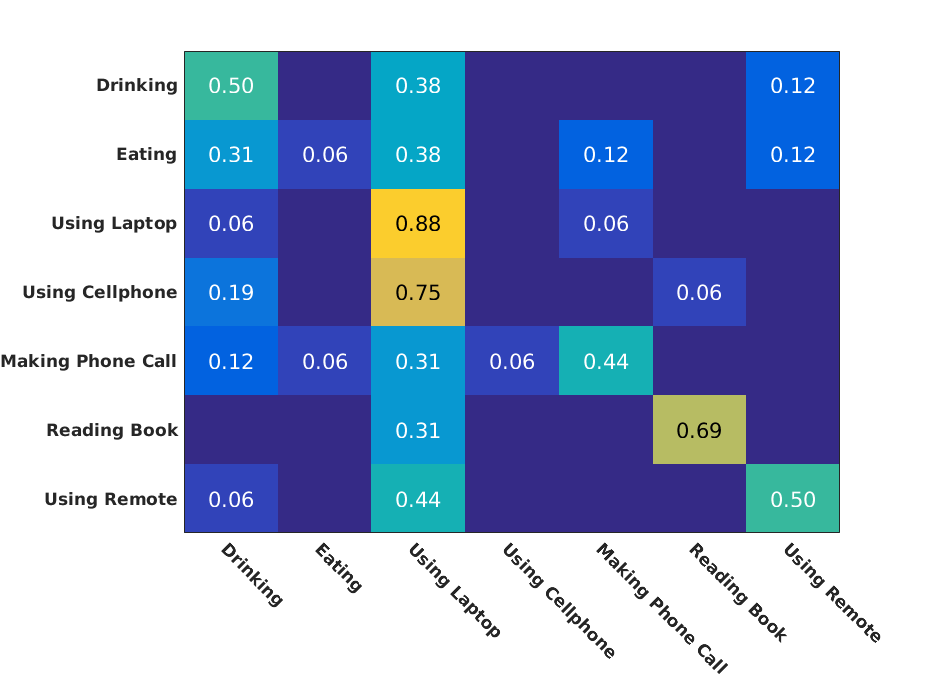}\\
(a)
\label{fig_first_case_orgbd}
\end{tabular}
\begin{tabular}[b]{c}
\includegraphics[width=0.8\textwidth]{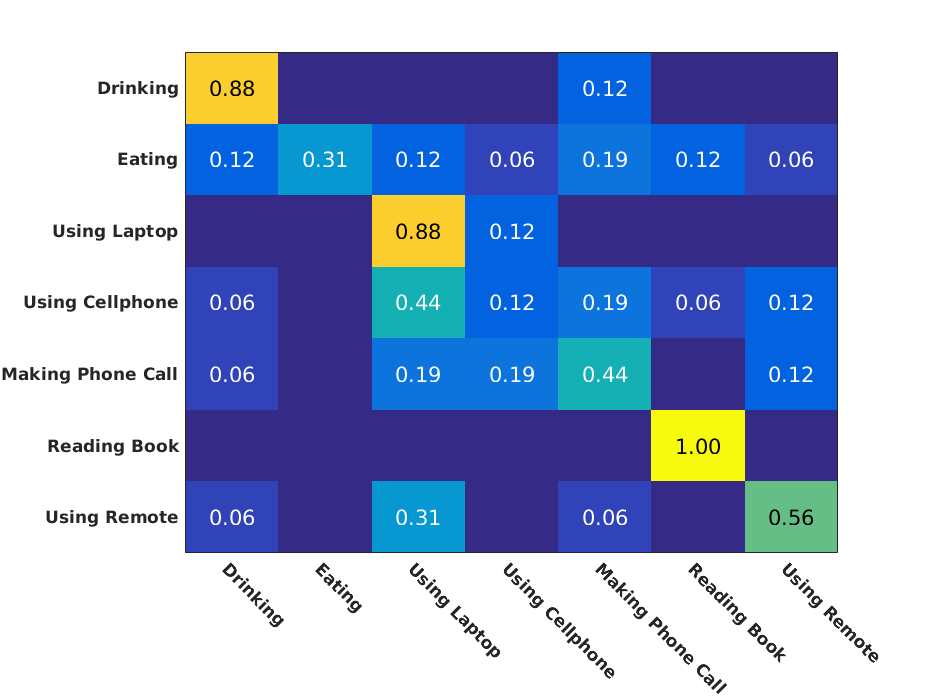}\\
(b)
\label{fig_second_case_orgbd}
\end{tabular}

\caption{Confusion matrices obtained for Dense Trajectories (a) and 2D Localized Trajectories (b) approaches (ORGBD).
}
\label{fig_orgbdconf}
\end{figure}

Despite its strong performances, 2D Localized trajectories action representation suffers from two limitations. First, 2D Localized Trajectories approach presents low performance when the motion amount is small. This attribute is inherited from Dense Trajectories approach and is clearly depicted in action classes such as \textit{Call Cellphone} in both MSR DailyActivity 3D and ORGBD as it is shown in Fig.~\ref{fig_msrconf} and Fig~\ref{fig_orgbdconf}, respectively, and \textit{Write on a Paper} in MSR DailyActivity 3D. Nonetheless, \textit{Sit Still} class achieves adequate performance with the use of 2D Localized Trajectories, since it is an action class with almost no motion.  

Second, 2D Localized Trajectories approach does not capture radial motion sufficiently. 
Action classes such as \textit{Playing the guitar} in MSRDailyActivity3D dataset include a notable amount of radial motion and the accuracy results were consequently low, as demonstrated in Fig.~\ref{fig_msrconf}a and Fig.~\ref{fig_msrconf}b. For that reason, as mentioned earlier, the proposed 3D Localized Trajectories presents as a good alternative to solve these two issues. Performance of the 3D Localized Trajectories are reported in the next section. 

\begin{table}[h!]
\renewcommand{\arraystretch}{1.3}
\caption{Mean accuracy of recognition (\%) on ORGBD dataset for Dense Trajectories and 2D Localized Trajectories approaches against literature in both Same and Cross Environment Settings}
\label{table_orgbd_se}
\centering
\begin{tabular}{|c||c||c|}
\hline
\multirow{2}{4em}{Method} & \multicolumn{2}{|c|}{Mean accuracy} \\
\cline{2-3}
& Same Env. & Cross Env.\\
\hline
\hline
Moving Pose \cite{zanfir2013moving} & 38.4\% & 28.5\%\\
\hline
Eigenjoints \cite{yang2012eigenjoints} & 49.1\% & 35.7\%\\
\hline
DSTIP \& DCSF \cite{xia2013spatio} & 61.7\% & 21.5\%\\
\hline
Skeleton \& LoP \cite{Wang12} & 66.0\% & 59.8\%\\
\hline
Pairwise joint distance \cite{yu2014discriminative} & 63.3\% & -- \\
\hline
Orderlet \cite{yu2014discriminative} & 71.4\% & -- \\
\hline
Dense Trajectories \cite{wang:2011:inria-00583818:1} & 64.3\% & 43.8\%\\
\hline
\textbf{2D Localized Trajectories} & \textbf{67.4\%} & \textbf{59.8\%}\\
\hline
\end{tabular}
\end{table}

\begin{table}[h!]
\renewcommand{\arraystretch}{1.3}
\caption{Mean accuracy of recognition (\%) on Watch-n-Patch in both kitchen and office settings for Dense Trajectories and 2D Localized Trajectories approaches.}
\label{table_wnp}
\centering
\begin{tabular}{|c||c|}
\hline
Method & Mean accuracy\\
\hline
\hline
Dense Trajectories (office)\cite{wang:2011:inria-00583818:1} & 68.8\%\\
\hline
Dense Trajectories (kitchen)\cite{wang:2011:inria-00583818:1} & 56.2\%\\
\hline
\textbf{2D Localized Trajectories (office)} & \textbf{71.1\%}\\
\hline
\textbf{2D Localized Trajectories (kitchen)} & \textbf{81.5\%}\\
\hline
\end{tabular}
\end{table}

\begin{table}[h!]
\renewcommand{\arraystretch}{1.3}
\caption{Mean accuracy of recognition (\%) of Dense Trajectories and 2D Localized Trajectories approaches on KARD dataset.}
\label{table_kard}
\centering
\begin{tabular}{|c||c|}
\hline
Method & Mean accuracy\\
\hline
\hline
JTMI \& LBP \& FLD \cite{ahmed2016joint} & 98.5\%\\
\hline
JTMI \& Gabor features \cite{tian2002evaluation} & 96.0\%\\
\hline
HOJ3D \cite{xia2012view} & 95.3\%\\
\hline
EigenJoints \cite{yang2012eigenjoints} & 96.2\%\\
\hline
Dense Trajectories \cite{wang:2011:inria-00583818:1} & 97.8\%\\
\hline
\textbf{2D Localized Trajectories} & \textbf{98.2\%}\\
\hline
\end{tabular}
\end{table}

\subsection{Performance of 3D Localized Trajectories}

The proposed 3D Localized trajectories approach was evaluated on MSRDailyActivity3D dataset. The results reported in Fig.~\ref{table_msr} show its superiority against Dense Trajectories and 2D Localized Trajectories. In fact, the accuracy of Dense Trajectories and 2D Localized Trajectories are improved by $1.9\%$ and $11.9\%$, respectively. 

The performance improvement happens mainly because of the inclusion of depth information in 3D trajectories. This helps in distinguishing actions which are performed radially with respect to the camera. The latter is particularly reflected in the confusion matrix of MSR DailyActivity 3D dataset in Fig.~\ref{fig_msrconf}, where actions 
like {\em play game} and {\em play guitar} are more effectively discriminated using 3D information.
The reported accuracies for the actions {\em play game} and {\em play guitar} are significantly improved. In particular, from $20\%$ and $20\%$ using Dense Trajectories and $40\%$ and $40\%$ using 2D Localized Trajectories, the accuracy climbed to $60\%$ and $70\%$ with the use of 3D Localized Trajectories, respectively.

These promising results highlight the potential of our first attempt to generalize Dense Trajectories to 3D and opens up new perspectives. Indeed, many components of this 3D concept can be reinforced to increase its effectiveness. For example, 3D trajectories are slightly more noisy than the Dense trajectories mainly because depth sensors introduce additional noise. This noise translated to a significant number of points belonging to the background which appeared to move radially, creating a lot of irrelevant 3D trajectories. 
Most importantly, the scene flow estimation is not optimal, since it relies on two different modalities which often appear to be misaligned. 
This fact is reflected in the performance of the 3D Trajectories (without locality), resulting in a notably lower accuracy than the Dense Trajectories, as demonstrated in Table~\ref{table_msr}. Nevertheless, the trajectory clustering around body joints is still able to remove a significant amount of noisy and irrelevant trajectories in 3D Localized Trajectories case.

\subsection{Global BoW vs. Local BoW}

To experimentally motivate the use of local BoWs, we compare the results obtained for 2D Localized trajectories using both a global BoW and a local BoWs. Hence, the experiments are conducted on the cross-environment scenario of the ORGBD dataset. The mean accuracy is notably lower compared to the 2D Localized Trajectories approach with Local BoW, reaching $53.6\%$ vs. $59.8\%$. The results suggest that trajectories clustering combined with local BoWs contribute significantly to the enhancement of the local discriminative power of the overall approach. They, also, suggest that the local encoding is more effective, since the codebooks are constructed using features which are specific to the motion of each body part.

\section{Conclusion}
\label{sec:conclusion}

In this paper, we proposed to solve two major shortcomings of the original Dense Trajectories approach using additional modalities provided by RGB-D cameras: the lack of locality information and the ineffectiveness in describing radial motion. Our contribution is two-fold. First, we enhanced the discriminative power and locality-awareness of Dense Trajectories by clustering them around human body joints. This method is coupled with the local Bag-of-Words concept, strengthening further the framework. Second, we constructed 3D Localized Trajectories for action recognition
. For this purpose, we used a) scene flow instead of optical flow for the generation of the 3D Trajectories and b) 4D extension of the originally used spatio-temporal descriptors. The reported results show the robustness of the two proposed representations in various challenging datasets. As future work, we intend to develop an automatic way of choosing the optimal parameters. 
In addition, we intend to estimate more reliable and robust to noise 3D Trajectories directly from point cloud data for the purposes of enhancing our current approach and extending it to view-invariant action recognition.  

\section*{Acknowledgment}

This work was funded by the European Union’s Horizon 2020 research and innovation project STARR under grant agreement No.689947, and by the National Research Fund (FNR), Luxembourg, under the project C15/IS/10415355/ 3DACT/Bj\"{o}rn Ottersten. Moreover, the experiments presented in this paper were carried out using the HPC facilities of the University of Luxembourg~\cite{VBCG_HPCS14} -- see \url{https://hpc.uni.lu}.

\end{document}